\pdfoutput=1

\documentclass[11pt]{article}

\usepackage[]{acl}

\usepackage{times}
\usepackage{latexsym}

\usepackage[T1]{fontenc}

\usepackage[utf8]{inputenc}
\usepackage[]{graphicx}
\usepackage{subcaption}
\usepackage{pdfpages}
\usepackage{enumitem}
\usepackage{cuted}

\usepackage{microtype}

\title{MedNgage: A Dataset for Understanding Engagement in Patient-Nurse Conversations}

\author{Yan Wang\textsuperscript{\rm 1}, Heidi Ann Scharf Donovan\textsuperscript{\rm 1}, Sabit Hassan\textsuperscript{\rm 2}, Malihe Alikhani\textsuperscript{\rm 2}\\
\textsuperscript{\rm 1} School of Nursing, \textsuperscript{\rm 2} School of Computing and Information\\
   University of Pittsburgh, Pittsburgh, PA\\
  \texttt {\{yaw75,donovanh,sabit.hassan,malihe\}@pitt.edu}}

\begin{document}
\maketitle
\begin{abstract}

Patients who effectively manage their symptoms often demonstrate higher levels of \textit{engagement} in conversations and interventions with healthcare practitioners. This engagement is multifaceted, encompassing cognitive and socio-affective dimensions. 
Consequently, it is crucial for AI systems to understand the engagement in natural conversations between patients and practitioners to better contribute toward patient care. In this paper, we present a novel dataset (MedNgage), which consists of patient-nurse conversations about cancer symptom management. We manually annotate the dataset with a novel framework of categories of patient engagement from two different angles, namely: i) socio-affective (\textbf{3.1K} spans), and ii) cognitive use of language (\textbf{1.8K} spans). 
Through statistical analysis of the data that is annotated using our framework, we show a positive correlation between patient symptom management outcomes and their engagement in conversations. Additionally, we demonstrate that pre-trained transformer models fine-tuned on our dataset can reliably predict engagement classes in patient-nurse conversations. Lastly, we use LIME \cite{ribeiro-etal-2016-trust} to analyze the underlying challenges of the tasks that state-of-the-art transformer models encounter. The de-identified data is available for research purposes upon request \footnote {\url{https://github.com/YanRayray/MedNgage}}.


\end{abstract}

\section{Introduction}
Due to the ease of use and efficiency of digital health interventions (DHIs) \cite{greaves2018appropriate}, we are witnessing a surge in online conversations between patients and healthcare providers. Literature suggests that actively engaged patients are more likely to obtain the full benefits of an intervention and exhibit better outcomes \cite{yardley2016understanding}. Therefore, it is critical to understand patients' \textit{engagement} in online conversations and extract insights to aid healthcare professionals in providing in-time support to the patients.

\begin{figure}
    \centering \includegraphics[width=0.95\linewidth]{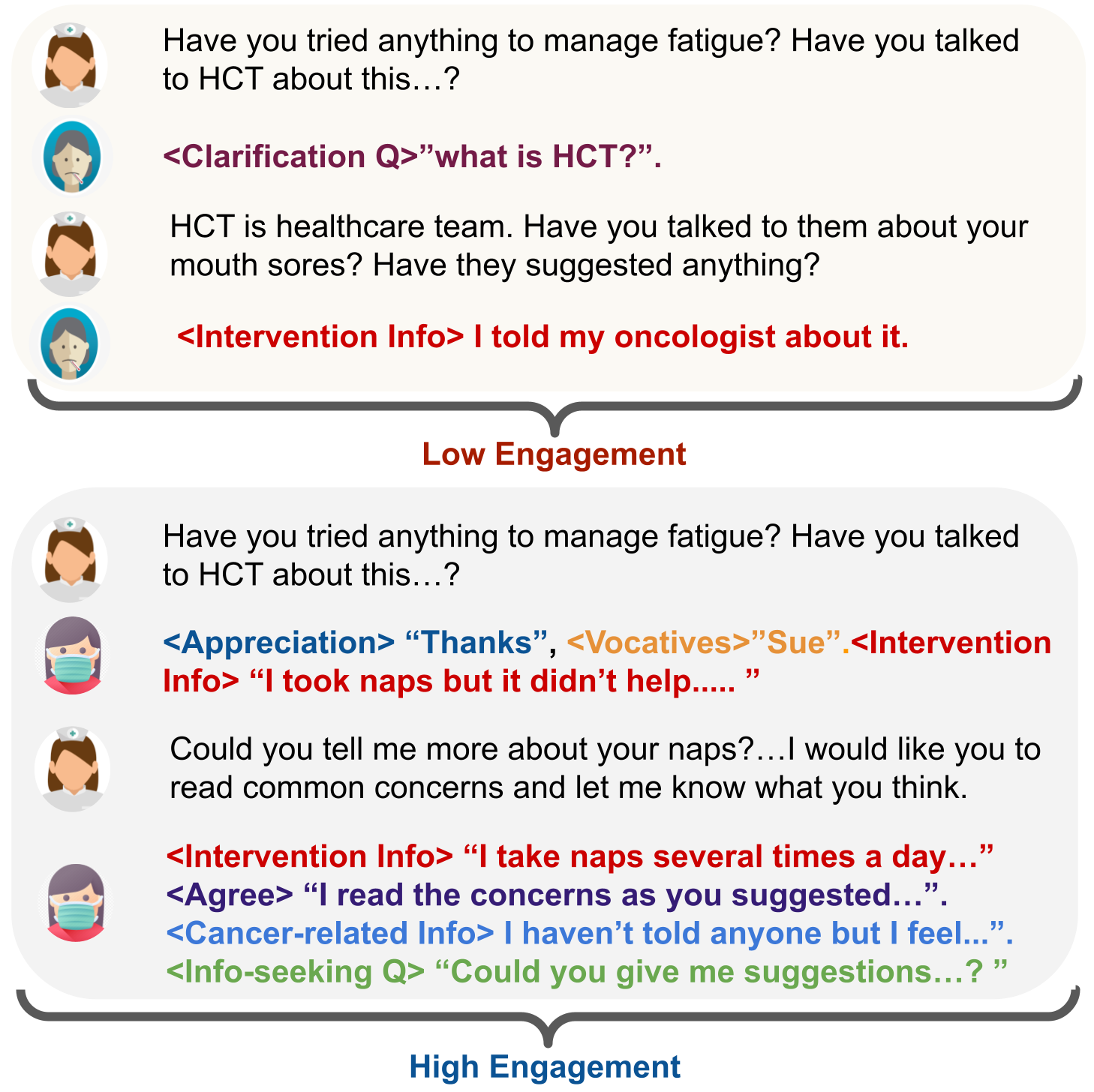}
    \caption{Our dataset contains patient-nurse conversations annotated with engagement derived from socio-affective and cognitive use of language. We hypothesize that patients who have high engagement tend to have better symptom control.}
    \label{fig:intro_fig}
\end{figure}

While previous research have studied modeling engagement in human-human and human-agent conversations \cite{reddy-etal-2021-modeling,sano-etal-2016-prediction,xu-etal-2020-discourse}, these works do not translate well to modeling engagement in patient-provider conversations. To bridge this gap, we introduce a novel resource called \textbf{MedNgage}, which consists of patient-private asynchronous message boards from an online intervention led by study nurses for symptom management. The dataset encompasses 2.1K turns between 68 patients and the nurses. We hypothesize that patient engagement (e.g., sharing cancer-related experiences, information-seeking question) can predict their symptom management outcomes. It is important to note that datasets containing patient-provider conversations are rarely found in the existing literature.

 



 
To effectively model engagement in our proposed dataset, we create a novel framework called Socio-Affective Cognitive Engagement (SACe), drawing inspiration from linguistic theories of discourse \cite{asher2003logics}, cognitive science of grounding in communication \cite{Clark1991GroundingIC}, and the model of social presence \cite{Swan2009ACA}. With SACe, we manually annotate the dataset, classifying patient engagement into eight classes of socio-affective engagement (e.g., personal information \cite{Swan2009ACA}) across 3.1K spans and seven classes of engagement derived from cognitive use of language (e.g., information-seeking question, clarification question \cite{asher2003logics}) across 1.8K spans within conversations. Additionally, we investigate the extent to which this framework can predict patient symptom management outcomes. Figure \ref{fig:intro_fig} showcases excerpts from our dataset, which have been annotated using the SACe framework.

\setlength{\tabcolsep}{3pt}

We conduct Spearman's rank correlation and Kruskal-Wallis tests to analyze the relationship between engagement classes and changes in patients' perceived symptom control. The tests provide empirical evidence that patients who have a higher level of engagement in certain classes gain more control over their symptoms. Thus, automated prediction of engagement from conversations can help practitioners to detect low patient engagement levels and identify effective intervention strategies so that they could tailor content for improved outcomes.
To facilitate this, we conduct a range of experiments evaluating the efficacy of both traditional machine learning (e.g., SVMs) and state-of-the-art transformer models in predicting engagement, followed by an analysis using LIME \cite{ribeiro-etal-2016-trust} to identify the challenges of the tasks. Thus, our contributions to this paper are:
%
\begin{itemize}[leftmargin=*]
\item  We present a novel framework for measuring engagement in patient-nurse conversations, as well as a novel dataset of engagement annotated with (i) \textbf{socio-affective} and (ii) \textbf{cognitive} use of language. The framework can be generalized to interactions in other healthcare scenarios that involve adopting healthy behaviors and improving clients' mental and physical well-being.
    \item With our statistical tests, we show a \textbf{positive correlation} between patient engagement level in conversations and perceived symptom control. 
   \item  We show transformer models can reliably predict engagement (F1 score $>$ \textbf{78}) when trained on our data to help practitioners identify patients with low engagement and intervene accordingly. We analyze classifier errors using LIME to provide insights into the challenges of the tasks.
\end{itemize}

\section{Framework}
Inspired by theories of discourse coherence, we explore linguistic and conceptual models of engagement across various disciplines. Through iterative collaboration with expert nurses and linguists during the initial content analysis phase, we have generated a set of classes that effectively capture the nuances of our data. Therefore, our categorization of engagement combines existing literature with data-driven adaptation to match the distinctive features of our dataset, MedNgage. We will discuss the theoretical underpinnings behind our proposed SACe framework and the classes of engagement for annotating our dataset. 





\paragraph{Socio-affective Use of Language (Category A)}
Interventions are social: in our online nurse-led symptom management intervention, the nurses worked to create a relationship with the patient based on trust, respect, and closeness and encouraged patients to talk about their feelings and concerns \cite{phillips2016nurses}. Thus, it is important for patients to feel affectively connected to the nurses and intervention. Adapted from social presence in the Community of Inquiry (CoI) framework \cite{Swan2009ACA}, for socio-affective engagement, we look for positive evidence of developing an affective connection (a trusting, close, and respectful therapeutic relationship) with the nurses. Appreciation, emotional expression, self-disclosure, greetings and salutations, and vocatives are examples of CoI categories that signal patient engagement in conversations. The mapping of CoI framework \cite{Swan2009ACA} to classes of socio-affective engagement is in Appendix \ref{appendixA}. However, CoI focused on online learning in higher education. Engagement indicators such as patients sharing their cancer experiences and interest in further communication are not included in this theory. Moreover, we excluded classes such as "humor" that do not appear commonly in our context. The eight socio-affective classes (Category A) are listed in Table \ref{tab:accents}.
\setlength{\tabcolsep}{4pt}
\renewcommand{\arraystretch}{1.1}
\begin{table*}
\small
\centering
\begin{tabular}{|l|l|r|}
\hline

\multicolumn{2}{|l|}{\textbf{Category A: Socio-affective use of language}} & \textbf{Count}\\
\hline
Appreciation & {"I really like the way you edited my goals.Thank you!" } & 343\\
\hline
Positive sentiment & {"I feel that all the strategies are good ones and THEY WORK!!" } & 357\\
\hline
Negative sentiment & {"I have now lost 2 replies! This will be my last try for tonight."} & 25\\
\hline
Cancer-related & {"I have caught myself cycling through emotional fatigue times where I’m}  &\\
experience & {completely frustrated with myself and life and want to give up and then} & 262\\
& {have some good news or an outing and everything is back to fine again.} &\\
\hline
Personal information & {"Yea. My daughter and son-in-law came to visit. We had  a great "} & 385\\ 
 & {Thanksgiving dinner.} & \\
\hline
Interest in communication& {"Please let me know if you need any other information."} & 161\\ 
\hline
Vocatives & {The patient addresses the study nurse by the name.} & 612\\
\hline
Greetings & {"Sincerely,", "Good morning", "With God's grace on us both,"} & 956\\ 
\hline
\multicolumn{2}{|l|}{\textit{Total Socio-affective spans}}& \textit{3101}\\
\hline
\hline
\multicolumn{2}{|l|}{\textbf{Category B: Cognitive use of language}} & \textbf{Count}\\
\hline

Intervention information & "It's hard to get up. I try to sleep late (until 8 or 8:30) and get in to work& 1224\\
&by 10, although I'd rather sleep until 10. I force myself to get up.” &\\ 
\hline
Information-seeking question & {"Are there other foods which make sores worse?"} & 177\\
\hline
Clarification question & {"Are you talking about a based on a 1 to 10 type of thing?"} & 14\\
\hline
Acknowledgment & {“I have received and read your email.”} & 50\\
\hline

Agreement & {“The care plan looks good. I don’t have anything to add.”} & 298\\
\hline

Disagreement & {“He won't read the material, his mind is made up.”} & 13\\ 
\hline
 & {"I think we can start with the sleep disturbance first. I am working on my} & \\
 Initiative taking &  eating now by myself, so that will take a while to see the affects of that, & 17\\
&  so let's move on to the next one, shall we."  & \\
\hline
\multicolumn{2}{|l|}{\textit{Total Cognitive engagement spans}}& \textit{1793}\\
\hline
\end{tabular}
\caption{Examples and total counts of patient engagement categories from the asynchronous message boards.} 
\label{tab:accents}
\end{table*}
\paragraph{Cognitive Use of Language (Category B)}

The primary goal of conversations between a patient and nurse in our dataset is to complete the intervention following the intervention protocols with some degree of individualization. To do so, it takes the nurse and the patient together to coordinate the content and plan of action on the message boards. They do it by building common ground (mutual understanding, assumption, and knowledge) and updating their common ground moment by moment \cite{Clark1991GroundingIC}. Therefore, we look for the positive evidence of patients coordinating the content and process of the intervention with the nurse within collaborative goals—acknowledgment and initiation of the relevant next turn (e.g., answering protocolized questions, clarification questions), based on \citet{Clark1991GroundingIC}. The mapping of grounding in communication framework \cite{Clark1991GroundingIC} to classes of cognitive use of language is in Appendix \ref{appendixA}. Moreover, we found \textit{taking initiative} to be an important patient engagement marker in our corpus that is not defined by \citet{Clark1991GroundingIC}. 
This class demonstrates patients taking the initiative on managing their symptoms without being prompted by the nurses. 

We have also studied discourse frameworks such as Segmented Discourse Representation Theory (SDRT) \cite{asher2003logics}. 
Discourse relation classes from SDRT such as information-seeking question, and clarification question overlap with \cite{Clark1991GroundingIC} and belong to our framework. The background class is closely related to Cancer-related experience in our data. Although there is an overlap between frameworks of engagement and discourse, engagement is a broader construct that captures a speaker's involvement in a conversation. While the discourse frameworks focus on logical and semantic relationships at Elementary Discourse Unit (EDU) level, engagement is observed across multiple EDUs in a conversation. As such, the remaining classes of SDRT and other discourse frameworks are not applicable directly. The final seven classes of cognitive use of language are listed in Table \ref{tab:accents} (Category B).



\section{Dataset}
\label{dataset}

In this section, we describe our data source, followed by our annotation protocol and a brief discussion about the patient characteristics in our dataset.  
\subsection{Data Source}
The data source for our work is asynchronous message boards between patients\footnote{Patient informed consent and approval of our institutional review board were obtained.} with recurrent ovarian cancer and study nurses. All interactions between the nurses and patients are captured verbatim. Patients interacted 1:1 with a nurse to go through the theory-guided intervention elements to develop individualized Symptom Care Plans for 3 target symptoms that they hoped to gain better control over (e.g., fatigue, pain, and nausea). A description of the intervention process is listed in Appendix \ref{appendixB}. 

One of the primary outcomes of the intervention is patient self-reported symptom controllability (i.e., an individual’s confidence in one's ability to control symptoms with medications and behaviors), which was assessed by a validated and reliable measure \textemdash 
Symptom Representation Questionnaire \cite{Donovan2008EvaluationOT}. First, the patients completed a 28-item symptom inventory (e.g., pain, fatigue, depression, nausea) and reported symptom severity at its worst in the past week from 0 (did not experience the symptom) to 10 (as bad as I can imagine). The patient then identified three target symptoms they would like to control better. The Symptom Controllability Scale (e.g., “I can do a lot to control this symptom”; five items) was used for each targeted symptom on a 0 (strongly disagree) to 4 (strongly agree) scale at the beginning and the end of the intervention.
\setlength{\tabcolsep}{5pt}

\begin{table}[]
    \small
    \centering
    
    \begin{tabular}{|c|c|c|c|}
    \hline
    
    \textbf{\# Patients} & \textbf{\# Patient} & \textbf{\# Nurse} & \textbf{\# Unique}\\ 
    &  \textbf{Turns} &  \textbf{Turns} &  \textbf{Tokens}\\    \hline

    68 & 1K & 1.1K & 29K\\
        \hline
        \hline
    
    \multicolumn{2}{|c|}{\textbf{\# Category A spans}} & \multicolumn{2}{c|}{\textbf{\# Category B spans}}\\
    \hline
    \multicolumn{2}{|c|}{3101} & \multicolumn{2}{c|}{1793}\\
    \hline
    \end{tabular}
    \caption{Overview of the annotated dataset. Category A refers to the socio-affective use of language and category B refers to the cognitive use of language.}
    \label{tab:data-stats}
\end{table}

\subsection{Manual Annotation}
Based on the theoretical framework presented earlier, two trained raters (one graduate and one undergraduate nursing student) independently begin by abstracting the sentences in the patients’ posts that reflect engagement by i) socio-affective and ii) cognitive use of language as meaning units. 
To determine the minimal meaningful units and their respective classes, the raters initiate the analysis by examining the first word of a patient's post. We then gradually expand the analysis, observing for points where the category of engagement changed. When a category shift is identified, the section analyzed up to that point is marked as a "span." If the current span does not fall under either category of engagement, it is skipped, and the analysis continues until the next category shift is detected. This iterative process is repeated until the entire post is classified into smaller sections or spans, each corresponding to a distinct category.
Examples from our annotation are listed in Table \ref{tab:accents}. An overview of the annotated dataset is provided in Table \ref{tab:data-stats}. 

\paragraph{Inter-rater Agreement} Code differences between the two raters were discussed and decided by the principal investigator of the intervention. 
A coding scheme with examples (Appendix \ref{appendixB}) is developed in an iterative manner to ensure inter-rater reliability. Inter-rater reliability is evaluated by Cohen $\kappa$ statistic across 131 turns by two annotators. Cohen $\kappa$ for are 0.86 and 0.87 for socio-affective and cognitive use of language, respectively.

\paragraph{Patient Characteristics} In the dataset we annotated, the mean age of the patients was 59.7 (SD = 9.5), ranging from 24 to 83. The majority (75\%) were married or living with a partner, and 51.5\% had a bachelor’s degree and above. Based on the Charlson Comorbidity Index (CCI), 45.6\% of the patients did not have any comorbidity, and 54.4\% had at least one comorbidity.

\setlength{\tabcolsep}{4pt}

\section{Analysis }
\label{analysis}
In this section, we first report the distribution of different engagement classes in our data. Then we illustrate the relationship between engagement and symptom controllability. The significance level for statistical analysis is p < 0.05.
\begin{table}[h]
\small
\centering

\begin{tabular}{|l|l|l|}

\hline

&	\textbf{Mean of }	& \textbf{Median of} \\
& \textbf{\textit{f} (SD)} & \textbf{\textit{f} (IQR)}\\
\hline
\hline
\multicolumn{3}{|l|}{\textbf{Category A: Socio-affective use of language}} \\
\hline
Positive sentiment & 3.63 (5.12)	& 2 (4.25) \\
\hline
Negative sentiment & 0.35 (0.99)	& 0 (0)	\\
\hline
Appreciation  & 3.96 (4.27) & 3 (5.25) \\
\hline
Cancer-related experience & 2.76 (2.97)	& 2 (3.25)\\
\hline
Personal information & 4.38 (3.84) & 3 (4)\\
\hline
Vocatives & 8.6 (9.11) & 6.5(10.5) \\
\hline
Interest in communication & 2.24 (2.73)	& 1 (3)	\\
\hline
Greetings  & 8.81(7.48) & 7 (8) \\
\hline
\hline
\multicolumn{3}{|l|}{\textbf{Category B: Cognitive use of language}} \\
\hline
Intervention information	& 12.41 (9.36)	& 10 (12.25)	\\
\hline
Acknowledgement	& 0.74 (1.19) & 0 (1) \\
\hline
Information-seeking question  & 2.26 (3) & 1 (3) \\
\hline
Clarification question & 0.21 (0.59) & 0 (0)\\
\hline
Agreement	& 3.38 (3.9) & 2 (6) \\
\hline
Disagreement	& 0.19 (0.4) & 0 (0) \\
\hline
Initiative taking & 0.22 (0.59)	& 0 (0)	\\
\hline
\end{tabular}
\caption{The mean and median of the frequency (\textit{f}) of each engagement category across 68 patients. }
\label{tab:descriptive}
\end{table}
\setlength{\tabcolsep}{4pt}

\setlength{\tabcolsep}{6pt}
\begin{table}[!h]
\small
\centering
\begin{tabular}{|l|l|}
\hline
\textbf{Engagement category}                           & \textbf{Freq-$\rho$ (p)} \\
\hline
\hline
\multicolumn{2}{|l|}{\textbf{Category A: Socio-affective use of language}}                      \\
\hline
Positive sentiment            & .35 (.008) \\
\hline
Appreciation                   & .25 (.007) \\
\hline
Cancer-related experience    & .34 (.01)   \\
\hline
Personal information                       & .34 (.01)  \\
\hline
Vocatives                  & .33 (.01)         \\
\hline
Interest in communication              & .32 (.01)    \\
\hline
Greetings                     & .27 (.04)      \\
\hline
\hline
\multicolumn{2}{|l|}{\textbf{Category B: Cognitive use of language}}\\
\hline
Intervention information  & .38 (.003)      \\
\hline
Information-seeking question              & .26 (.05)     \\
\hline
Agreement                                          & .34 (.009)          \\
\hline
Initiative taking              & .26 (.047)\\
\hline
\end{tabular}
\caption{Correlations between controllability score changes and the frequency of each engagement class. Insignificant results are not reported (p > 0.05).}
\label{tab:correlation}
\end{table}

\subsection{Distribution of Engagement}
The most common socio-affective engagement classes are Greetings (mean = 8.81, SD = 7.48) and Vocatives (mean = 8.6, SD = 9.11). The least common one is expressing Negative sentiment (mean = 0.35, SD = 0.99). 

The most commonly used class in cognitive use of language is Intervention information (mean = 12.41, SD = 9.36). The least common ones are Disagreement (mean = 0.19, SD = 0.4), Clarification question (mean = 0.21, SD = 0.59 ), and Initiative taking (mean = 0.22, SD = 0.59).
Table \ref{tab:descriptive} lists the mean and median of the frequency of each engagement class. 

\begin{figure*}[h]
\begin{centering}
\begin{subfigure}{.314\textwidth}
  \centering
  \includegraphics[width=.95\linewidth,height=4cm,keepaspectratio]{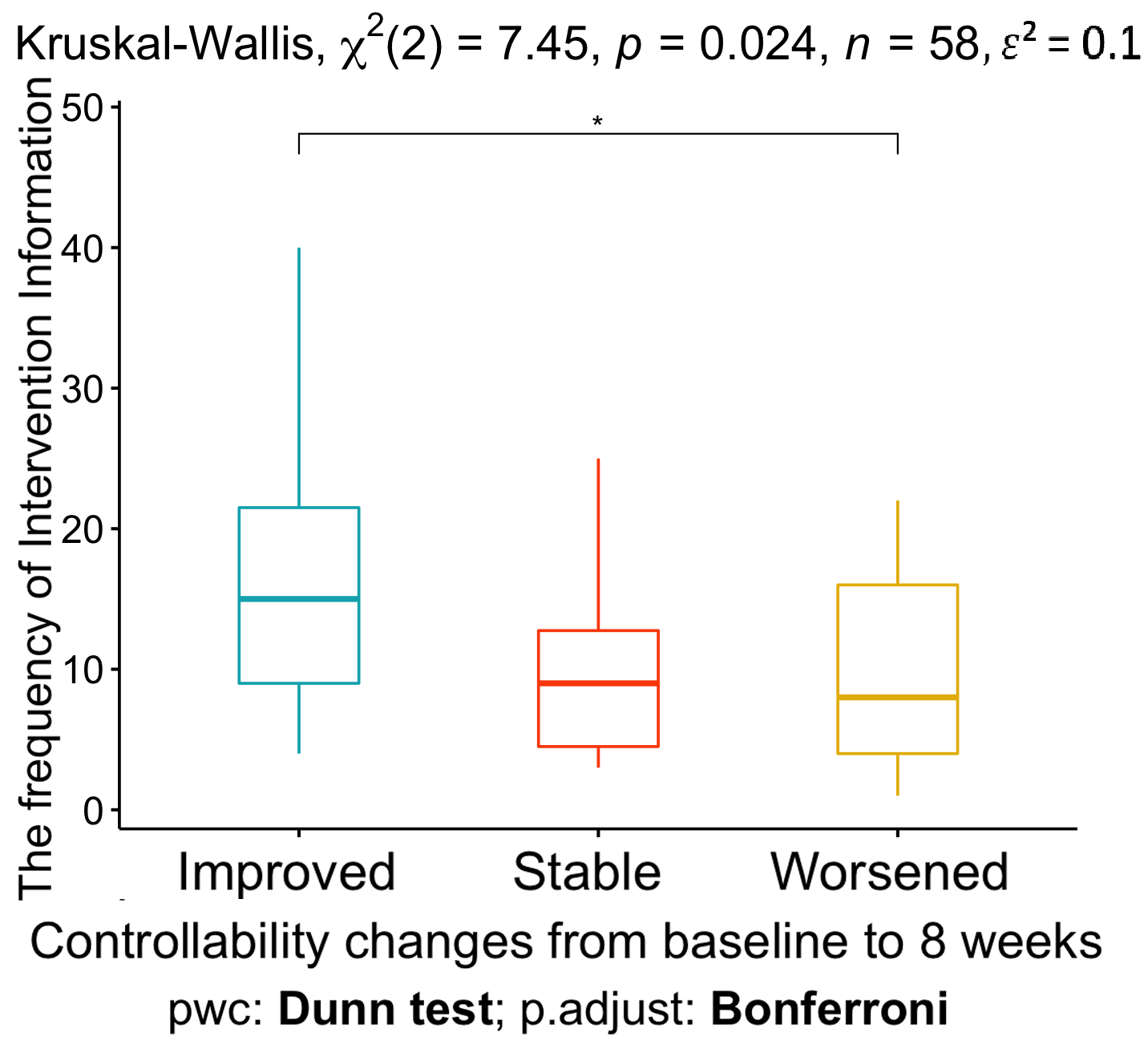}
  \caption{
  \centering
  Intervention Information
  }
  \label{fig:sfig1a}
\end{subfigure}
\begin{subfigure}{.314\textwidth}
  \centering
  \includegraphics[width=.95\linewidth,height=4cm,keepaspectratio]{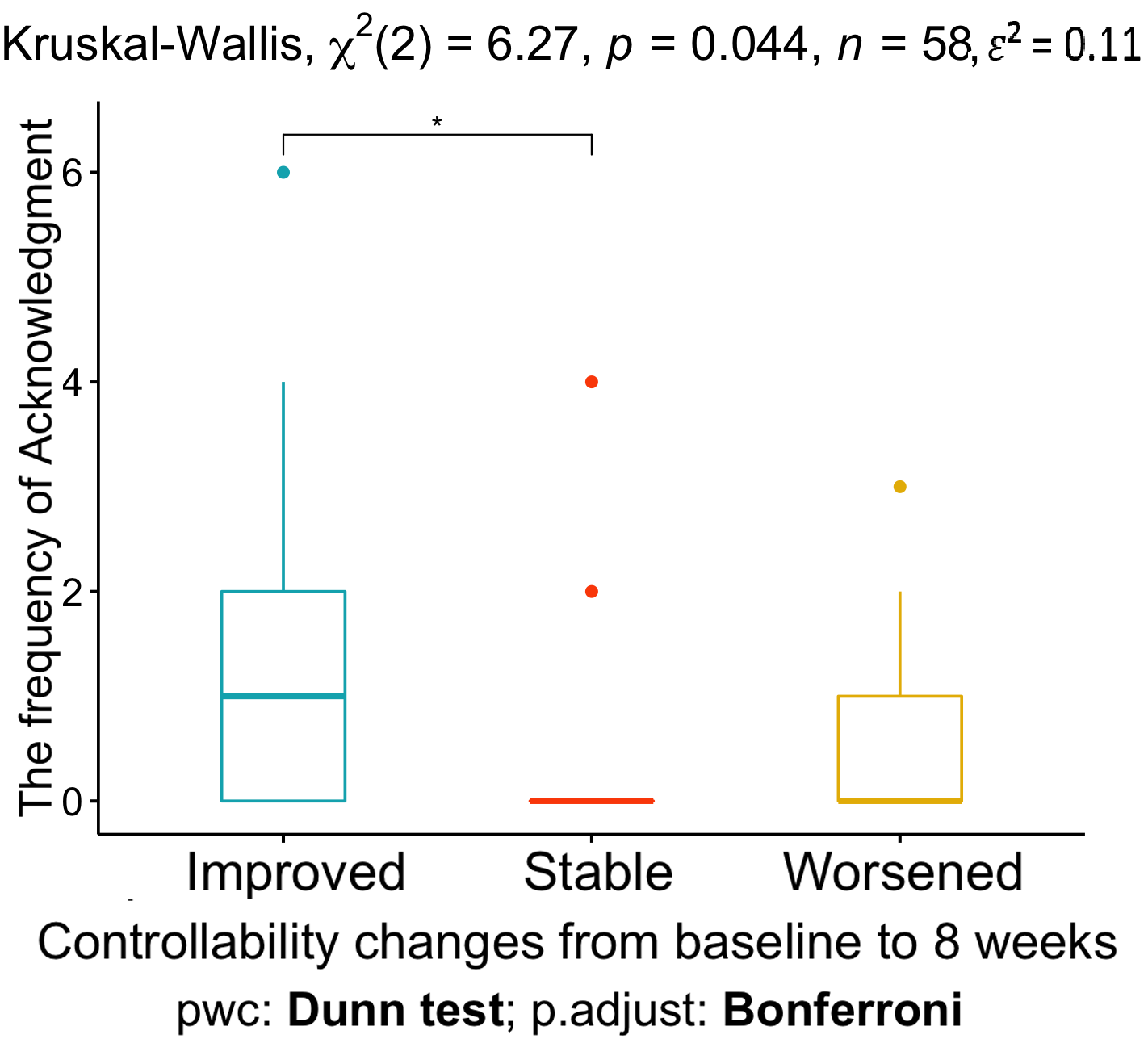}
  \caption{Acknowledgment}
  \label{fig:sfig1b}
\end{subfigure}%
\begin{subfigure}{.314\textwidth}
  \centering
  \includegraphics[width=.95\linewidth,height=4cm,keepaspectratio]{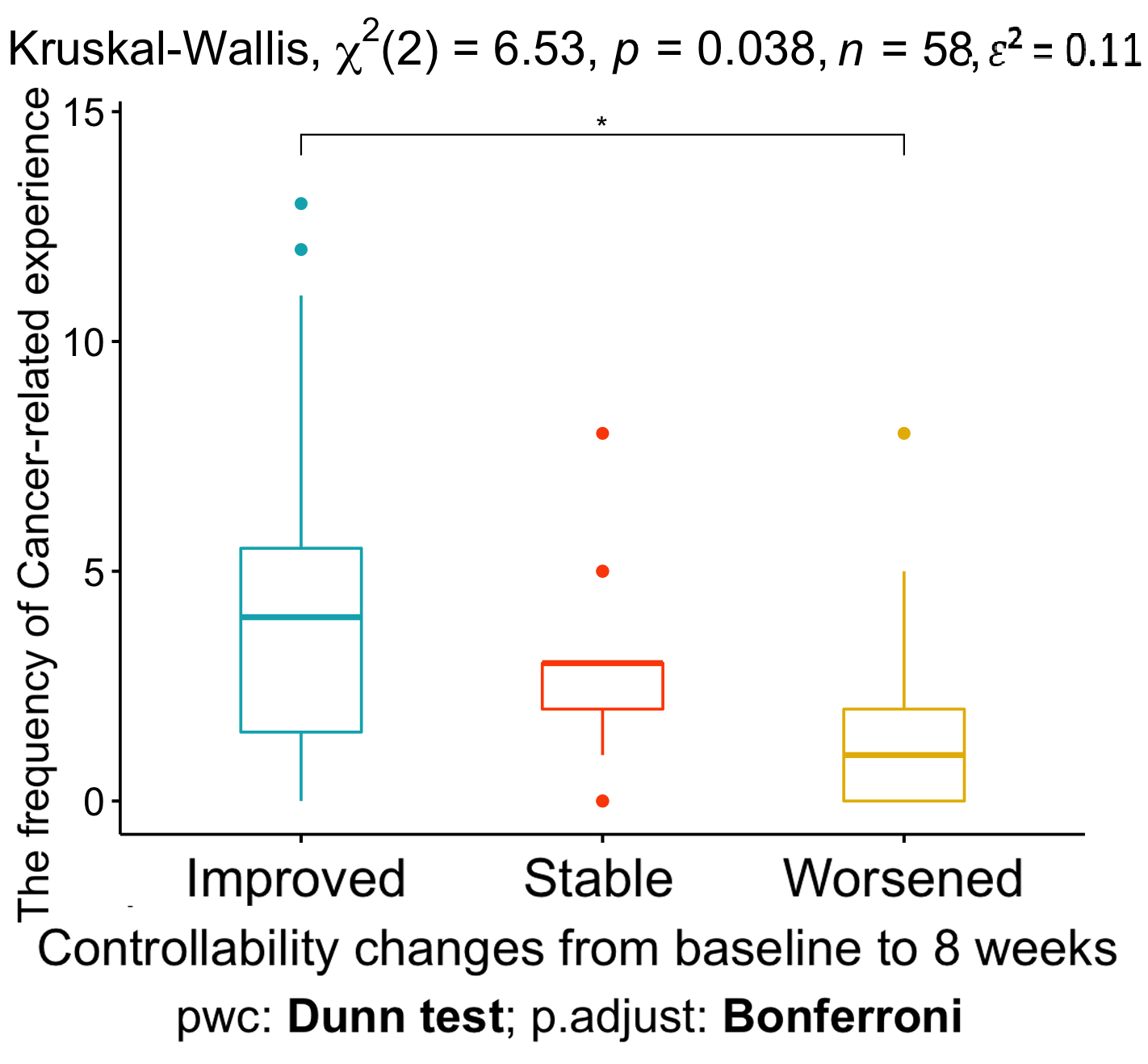}
  \caption{Cancer-related experience}
  \label{fig:sfig1c}
\end{subfigure}
\begin{subfigure}{.314\textwidth}
  \centering
  \includegraphics[width=.95\linewidth,height=4cm,keepaspectratio]{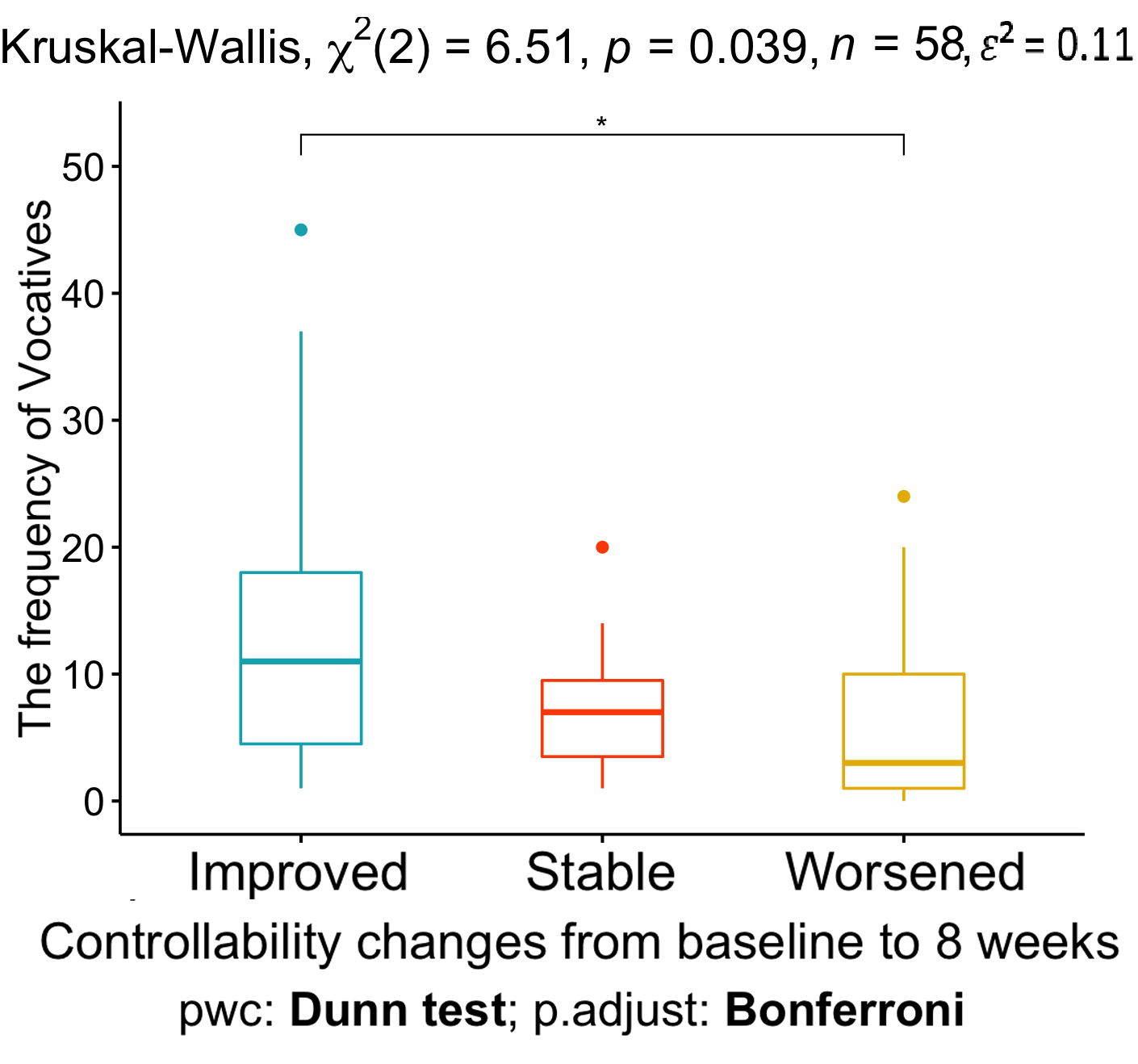}
  \caption{Vocatives}
  \label{fig:sfig1d}
\end{subfigure}
\begin{subfigure}{.314\textwidth}
  \centering
  \includegraphics[width=.95\linewidth,height=4cm,keepaspectratio]{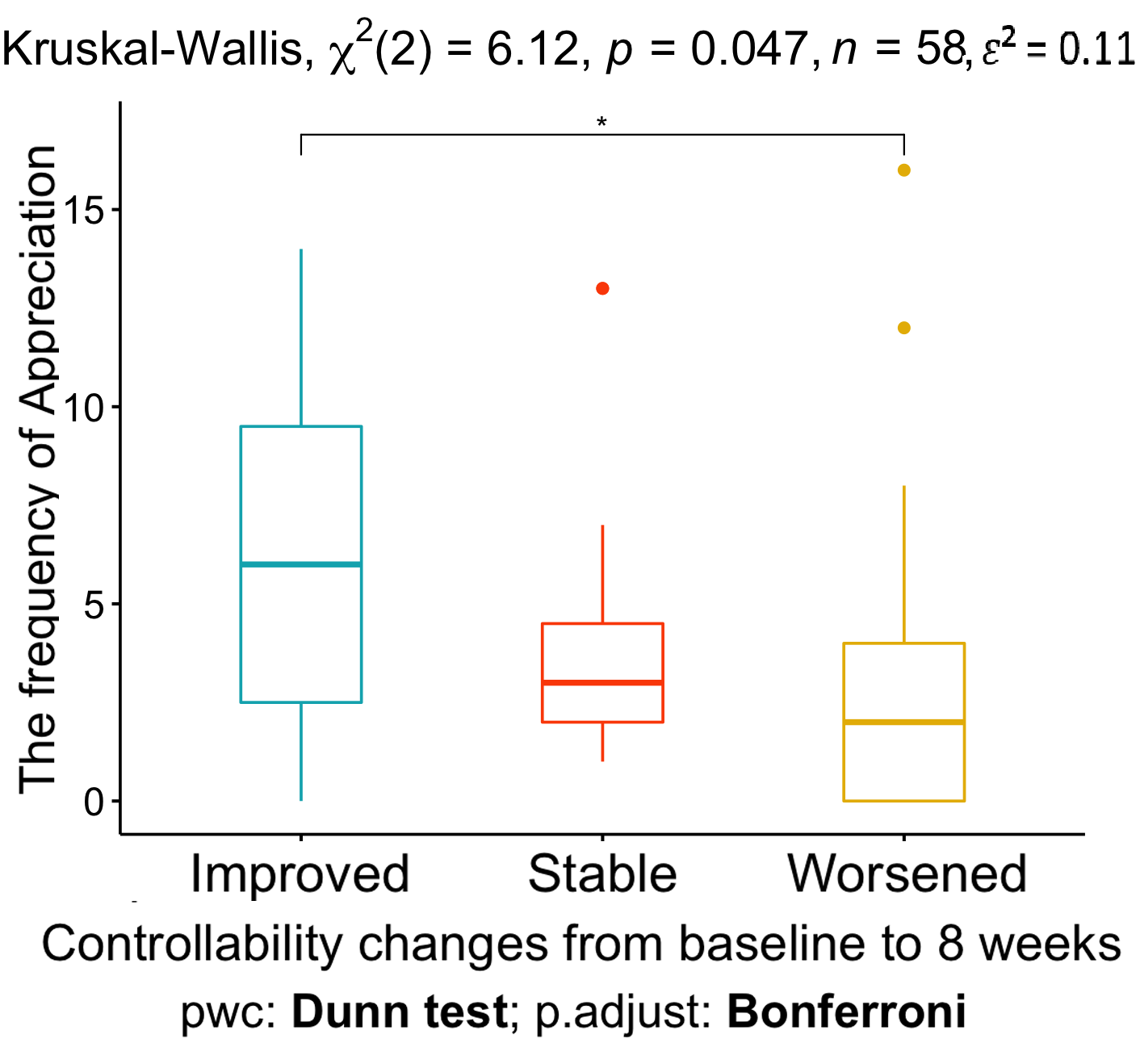}
  \caption{Appreciation}
  \label{fig:sfig1e}
\end{subfigure}
\caption{Kruskal-Wallis tests and post-hoc Dunn's tests on the frequencies of engagement classes between patients with an improved vs stable vs worsened sense of control over their symptoms. * indicates statistical significance (p $\leq$ 0.05).}
\label{fig:fig}
\end{centering}
\end{figure*}

\subsection{Symptom Controllability and Engagement}

Of 68 patients, 58 reported symptom controllability scores at baseline and at the end of the intervention (8 weeks). The average controllability score change is 0.21 (SD = 0.61), ranging from -1.85 to 1.73. Based on symptom controllability score change, we divided patients into three groups: (1) improved (n = 23) with changes greater than +0.3 SD (> 0.389), (2) stable (n = 14) with changes within +/-0.3 SD (0.029-0.389), and (3) worsened (n = 21) with changes greater than -0.3 SD (0.029).


Spearman’s rank correlation was computed to assess the correlations between controllability and engagement. Most classes' frequencies have a significant positive relationship with changes in patient controllability scores from baseline to 8 weeks (0.2<$\rho$< 0.4).  Table \ref{tab:correlation} shows the correlations between patient controllability scores and the frequency of each class. Kruskal-Wallis tests \cite{kruskal1952use} were used to examine the differences in the frequency of engagement classes among patients with an improved vs. stable vs. worsened sense of control over their symptoms. When a significant difference was detected, a post-hoc Dunn’s test \cite{Dunn1961MultipleCA} with Bonferroni adjustment \cite{Napierala2014WhatIT} for multiple pairwise comparisons was applied to distinguish the significant and non-significant pairs. We used $\epsilon\textsuperscript{2}$ to calculate the effect sizes for the Kruskal-Wallis tests and biserial correlation r for post-hoc Dunn’s tests. Although the sample size is smaller, we were able to detect a moderate to large effect size.

We discovered variations in the frequency of specific engagement classes between patients who experienced improved control and those with stable or worsening control. Regarding cognitive us of language, patients who reported improved perceived control over symptoms provided information toward intervention (i.e., content, process, and technical issues) more often (Mdn = 17) than those who reported worsened control (Mdn = 8), with an effect size of 0.4 (Figure \ref{fig:sfig1a}). Individuals who reported improved control acknowledged the nurse’s contributions more frequently (Mdn = 1) than those who had a stable sense of control (Mdn = 0), with an effect size of 0.41 (Figure \ref{fig:sfig1b}). Although marginally significant (P = 0.053), patients who reported improved control agreed with the nurse or agreed to do the intervention activities more frequently (Mdn = 4) than those who reported worsened control (Mdn = 1), with an effect size of 0.36. 

In terms of socio-affective engagement classes, individuals who reported improved controllability shared cancer-related experience (e.g., cancer story, vulnerability) more often (Mdn = 4) than those who reported worsened symptom control (Mdn = 1), with an effect size of 0.38 (Figure \ref{fig:sfig1c}). Patients who reported an improved sense of control addressed nurses by their names more frequently (Mdn = 11) than those who reported worsened symptom control (Mdn = 3), with an effect size of 0.39 (Figure \ref{fig:sfig1d}). Moreover, patients who reported improved controllability appreciated and recognized the nurses’ contributions significantly more often (Mdn = 6) than those who reported worsened symptom control (Mdn = 2) (Figure \ref{fig:sfig1e}). 
Although marginally significant (P = 0.053), patients who appeared to report an improved sense of control expressed positive sentiment toward intervention more frequently (Mdn = 4) than those who reported worsened symptom control (Mdn = 1), with an effect size of 0.37.


\section{Experiments}
In order to assess the feasibility of predicting classes of engagement, we train traditional SVMs and fine-tuned pre-trained transformer models on our dataset. 

\subsection{Dataset for experiments}
For our experiments, we isolate the spans and use 80\% of the data for training, 10\% for validation, and 10\% for testing. All numbers are reported on the test set. Due to the relatively small number of instances, we leave out the low-frequency classes ("Disagreement" and "Clarification questions"), yielding eight classes of socio-affective (Category A) and five classes of cognitive (Category B) use of language. 
\begin{figure*}[h]
    \centering
    \includegraphics[width=.95\linewidth]{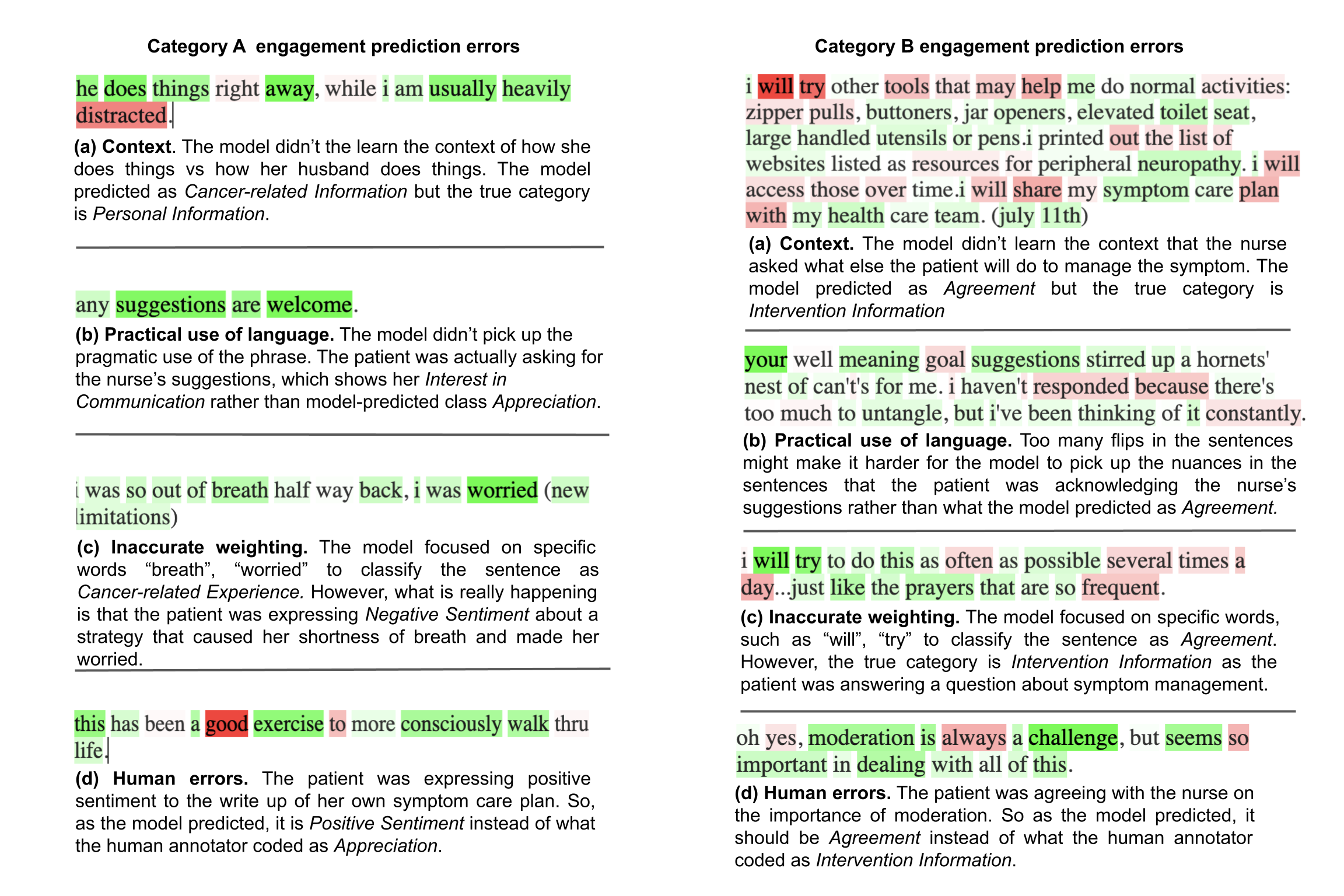}
    \caption{Examples of LIME output on classifier errors, along with the error type. 
    The socio-affective (Category A) engagement prediction emphasizes words that demonstrate colloquial usage while the cognitive use of language prediction (Category B) emphasizes words with the patient's goal. 
    Green highlighted words have positive weights and contribute to the classifier predicting a certain class, while red highlights have negative weights and reduce the likelihood of that class. The darker the color, the greater the impact of the word on the prediction. }
    \label{fig:LIME}
\end{figure*}
\setlength{\tabcolsep}{7pt}
\begin{table}[!h]
    \small
    \centering
    \begin{tabular}{|l|c|c|c|c|c|}

    \hline
    \multicolumn{4}{c}{\textbf{Category A: Socio-affective use of language}}\\
        \hline

        \textbf{Model} & \textbf{P} & \textbf{R} & \textbf{F1}\\
    \hline
    \hline
     SVM-W & 81.1 (0.0) & 39.8 (0.0) & 40.7 (0.0)\\
     SVM-C & \textbf{84.7 (0.0)} & 59.7 (0.0) & 65.6 (0.0)\\
     Bio-BERT &	75.4 (2.1) & 78.0 (1.1) & 76.4 (1.7)\\
     BERT & 75.1 (1.1) & 76.6 (0.5) & 75.6 (0.6)\\
     XLNet & 77.4 (1.9) & 78.7 (1.6) & 77.7 (1.4)\\
     RoBerta & 78.6 (1.3) & \textbf{79.6 (1.2)} & \textbf{78.8 (1.0)}\\
    
    \hline
    \hline
        \multicolumn{4}{c}{\textbf{Category B: Cognitive use of language}}\\

    \hline
        \hline

    \textbf{Model} & \textbf{P} & \textbf{R} & \textbf{F1}\\
    \hline
    \hline
     SVM-W & 62.1 (0.0)	& 58.9 (0.0) & 57.6 (0.0)\\
     SVM-C	& 72.1 (0.0)	& 73.5	(0.0) & 72.5 (0.0)\\
    Bio-BERT  &	\textbf{78.0 (2.1)}	& \textbf{79.3 (0.4)} & \textbf{78.5 (1.3)}\\
    BERT	& 77.5 (0.9) & 78.3 (0.6) & 77.6 (0.6)\\
    XLNet & 61.9 (0.8) & 63.9 (2.3) & 62.8 (0.7)\\
    RoBerta & 61.5 (2.3) & 66.9 (0.2) & 63.8 (1.3)\\
    \hline
    \end{tabular}
         \caption{Results of predicting engagement. Means across three runs are reported along with standard deviation in parenthesis. SVM-W and SVM-C refer to SVMs trained with word and character n-grams. RoBERTa achieves the best performance for Category A and Bio-BERT achieves the best performance for Category B.}

    \label{tab:results}
\end{table}

\subsection{Classification}
\paragraph{SVMs} As baselines, we train SVMs on word bigrams and character n-grams (2-5), weighted by term frequency-inverse document frequency (tf-idf) using scikit-learn's\footnote{\url{https://scikit-learn.org/stable/modules/generated/sklearn.svm.LinearSVC}} default parameters.
\paragraph{Pre-trained transformers}

We fine-tune four pretrained transformers: i) bert-base-cased \cite{Devlin2019BERTPO}, ii) BioClinicalBERT \cite{Lee2019BioBERT}, iii) RoBERTa-base \cite{liu2019roberta} and iv) XLNet-base-cased \cite{xlnet}. All the transformer models are trained for 3 epochs under the same settings: learning rate of 8e-8, batch size of 16, and a maximum length of 200 tokens.

\paragraph{Experiment results} The experiment results are presented in Table \ref{tab:results}. We report macro-averaged precision, recall, and F1 scores. Each experiment is repeated three times and the mean results, along with the standard deviation are reported. Since SVMs are deterministic, the standard deviation is 0. We observe that fine-tuned transformer models can reliably predict engagement in our dataset with mean F1 scores of \textbf{78.8} and \textbf{78.5} respectively for socio-affective and cognitive use of language.

\label{experiments}


\subsection{Error Analysis}
Despite the promising results achieved by transformer models, it is important to assess the limitations of these models due to the sensitivity of patient conversations. Thus, we manually annotate all instances where the best model (BioClinicalBERT for cognitive, and RoBerta for socio-affective engagement) make errors. To aid our analysis, LIME \cite{ribeiro-etal-2016-trust} is used to identify the words and phrases that support the models’ selection of a particular class. LIME determines the contribution score of specific words on the prediction of a classifier by generating variations of an input sentence by randomly removing a word and observing changes in the prediction. The "contribution score" represents how much weight the word had in the original prediction by the model.

\begin{figure}
    \centering
    \includegraphics[width=.90\linewidth]{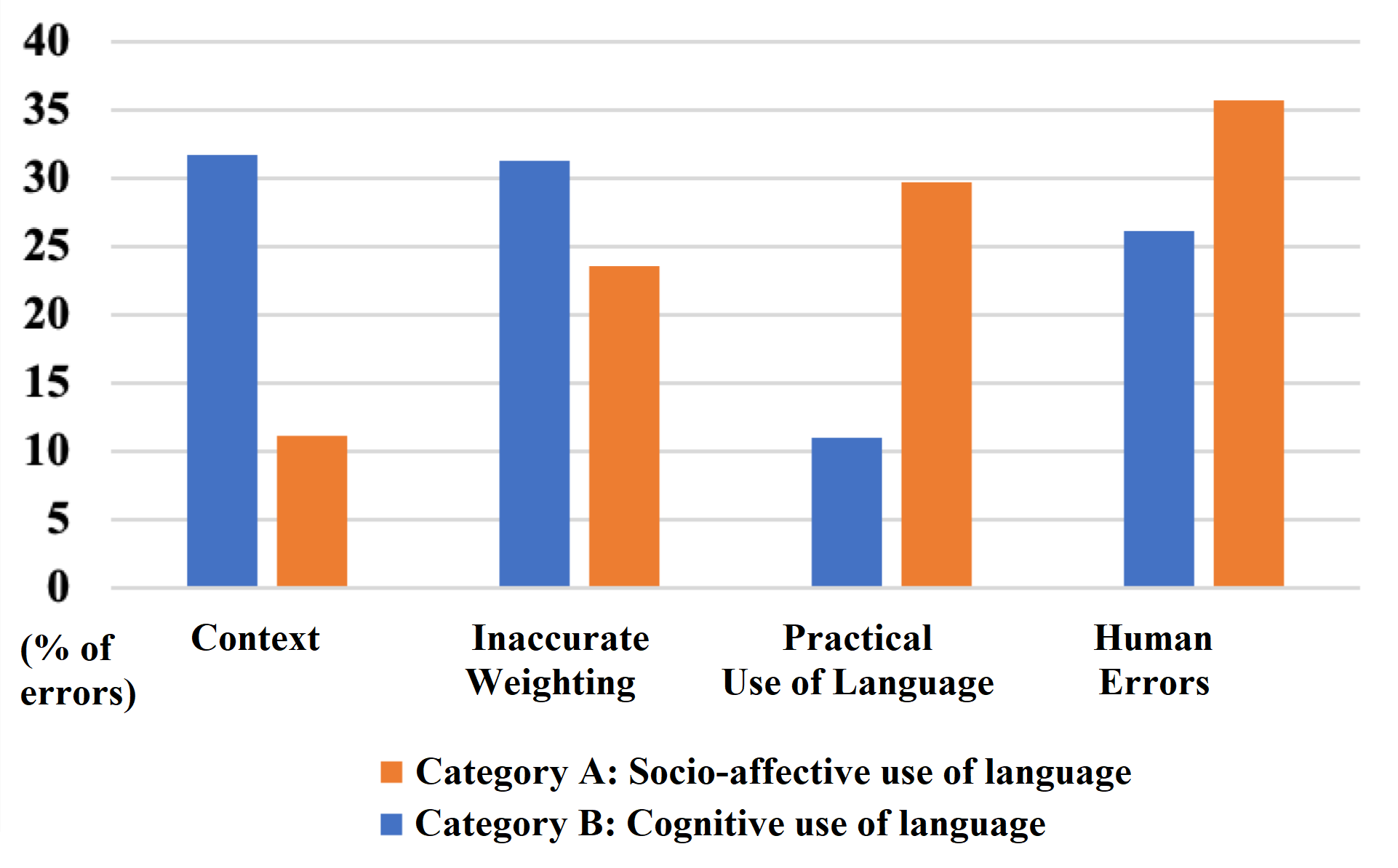}
    \caption{Weighted percentages of errors based on LIME. The socio-affective model tends to make pragmatic errors, while the cognitive model tends to make context errors.}
    
    \label{fig:error-dist}
\end{figure}

Observing the output of LIME, we identify four main types of errors: i) missing context, ii) inaccurate weighting of words; iii) practical use of language; and iv) human error in the annotation. Figure \ref{fig:LIME} shows the examples of errors and Figure \ref{fig:error-dist} shows the weighted percentages of errors based on the count of classes for the two tasks. 

\paragraph{Errors for Category A:}The most frequent error types for socio-affective engagement, apart from human error, are pragmatic failures and inaccurate weighting, which account for 53\% of the errors after being adjusted for the count of the socio-affective engagement classes. Due to the failure in the practical use of language (10\% of the errors evaluated after the adjustment), the model tends to make mistakes between the classes of Appreciation and Positive sentiment. Similarly, the human error rate is also the highest between these two classes (10\% of the errors analyzed after the adjustment).

\paragraph{Errors for Category B:} Context mistakes and inaccurate weighting account for around 63\% of the errors in the prediction of cognitive use of language after being adjusted for the count of the engagement classes. Due to these two types of errors, the model commits a high rate of errors in the classes of Intervention information and Agreement (15\% of the errors examined after the adjustment for each type of error). Of these two categories, human mistake rates also account for 15\% of the errors. Due to inaccurate weighting, the model made errors in differentiating the classes of Information-seeking question and Agreement (16\% of the errors after adjustment). 

We find a large difference in errors of practical use of language and context errors between cognitive and socio-affective tasks. We think this is because when predicting socio-affective engagement, the model needs to consider more pragmatic factors (e.g., situational context, the individuals' mental states) than cognitive use of language. For example, in Figure \ref{fig:LIME}b which shows LIME output, we observe that the model focuses on the word “welcome”, which is an indicator of “Appreciation”, but in practical usage, it was used as interest in communication. On the other hand, detecting cognitive use of language relies more on tone and structure, leading to context errors in Figure \ref{fig:LIME}a.




\section{Implications on DHIs}
\label{implications}
By examining the relationship between patients’ engagement and their symptom controllability, our study provides empirical evidence of the potential pathways of patient learning and behavior change to gain better control over symptoms through meaningful engagement. The classification models presented in this paper can be used to track patient engagement based on patient-provider interactions on asynchronous message boards. The models could be used as complementary tools to augment clinicians’ capabilities to identify patients with a low engagement level so that they can focus their energy and time (i.e., tailoring content and communication accordingly) to enhance engagement and help those who struggle the most to obtain the maximum benefits of a given DHI. Based on predicted engagement markers, or lack thereof, the provider can tailor the intervention content (e.g., suggest more strategies based on professional experience, segment a series of questions into a few posts, provide timely emotional support) to achieve common ground and cultivate an affective connection with the patient. For example, nurses can encourage the patients to i) provide relevant information to complete intervention activities and tasks (e.g., describing their symptom beliefs), and ii) share their feelings, cancer treatment, and symptom experiences. For patients participating in DHIs to achieve desired outcomes, it is also important to feel close to the provider, for example, addressing the provider by name, acknowledging the provider's message, and recognizing their contributions to the care and support.

Our dataset, MedNgage contains rich narratives of various symptom experiences and management processes from women who were fighting advanced cancer in an online cognitive behavior intervention. The Representational Approach \cite{Donovan2007AnUO} underlying the intervention is disease-agnostic and designed to help clients understand how their disease representations relate to behaviors. Therefore, the annotation system and model developed in this study can be applied to various healthcare scenarios involving the promotion of healthy behaviors and the enhancement of clients' mental and physical well-being. The corpus includes comprehensive representations of patient symptoms, concerns, and the obstacles they faced while seeking the best symptom management. This presents an excellent opportunity for various clinical NLP tasks, such as developing a dialog system, optimizing patient self-report/expression review, generating text summaries of patient conversations, and customizing automated relevant responses to improve patient engagement, similar to works on summarization \cite{xu-etal-2020-discourse} and style transfer \cite{Atwell2022APPDIAAD}. For example, the system can use patient-specific language or generate text that encourages patients to ask questions or express their concerns. 

\section{Related Work}
\label{related}
Engagement in human-human conversation has been studied from both socio-affective and cognitive aspects, resulting in numerous frameworks and theories for modeling engagement. One of the inspirations behind our work, the Community of Inquiry framework \cite{Swan2009ACA}, which builds on the educational philosophy and practice of \citet{DeweyMyPC}, models conversations in the online learning environment in terms of social presence, cognitive presence, and teaching presence. \citet{Rourke1999AssessingSP}, proposes a community of inquiry framework that synthesizes pedagogical principles with the dynamics of computer conferencing, focusing on social presence. \citet{Clark1991GroundingIC}, the inspiration behind the cognitive aspect of engagement in our work, studies how grounding in conversation is shaped by \textit{purpose} and \textit{medium} of the conversation. The authors highlight different grounding references that we adapt to our work. 


Previous research has investigated engagement in medical interactions, including conversations. Studies have specifically examined the cognitive engagement of individuals with schizophrenia in conversations, with a focus on medication adherence \cite{McCabe2013SharedUI, Howes2012PredictingAT}. In contrast, our study employs various methods to identify and predict patient engagement in a complex intervention aimed at modifying multiple health behaviors, including diet, physical activity, relaxation, and medication adherence. Further, we investigate the impact of engagement on patient symptom outcomes, providing practical strategies for nurses and other practitioners to effectively engage patients and deliver optimal care.

Other studies have explored engagement in different contexts such as political argument settings \cite{Shugars2019WhyKA}, conversations around terrorist attacks \cite{Chiluwa2016OnTA}, socio-affective aspects of conversations such as emotion \cite{Yu2004DetectingUE}, student engagement in online discussion forums \cite{Liu2018InvestigatingRB}, cognitive engagement in MOOC forums \cite{Wen2014LinguisticRO}, real-time engagement in reducing binge drinking through intervention text messages \cite{irvine2017real}, user engagement in online health communities \cite{Wang2020PredictingUC}. However, none of these studies specifically consider the dynamics of socio-affective and cognitive use of language in online conversations between patients and healthcare providers. Our work is the first to develop a dataset with aggregated annotations and models for predicting engagement in patient-provider scenarios.



%

\section{Conclusions}
\label{conclusion} 
We have developed a framework \textbf{(SACe)} that effectively captures patient engagement in patient-nurse conversations. Through the analysis of a unique dataset \textbf{(MedNgage)} consisting of online patient-nurse conversations, we have identified eight classes for socio-affective engagement and seven classes of engagement for cognitive use of language. These findings provide valuable insights for behavioral scientists to understand and monitor patients' engagement in healthcare interactions. The approach could be applied in other fields, such as online education, where engagement plays a crucial role. 
Our analysis confirms that higher levels of engagement, including the increased coordination and emotional connections between patients and healthcare practitioners during the intervention, result in improved symptom control.
Additionally, we have demonstrated that fine-tuned transformer models can reliably predict fine-grained engagement in conversations so that practitioners can adjust their communication style and tailor strategies promptly, to promote patient engagement for improved outcomes. Our analysis of model output using LIME has shown the challenges that state-of-the-art transformer models encounter for the two tasks, which can be used to improve these models for similar tasks in the future. We expect our system could also aid subsequent text generation tasks, such as summarization of patient conversations, and tailoring automated responses in clinical settings.

\section*{Limitations}
\label{limitation}
While this dataset is unique and pioneering, its size is limited, and it involves specific patients. To enhance the generalizability of the findings, a larger dataset may be required. Similarly, although our framework is innovative, we anticipate the development of more comprehensive and informative annotation protocols in the future. For instance, we observed a higher frequency of the "Intervention information" category within cognitive engagement, likely because the intervention predominantly follows a Q (nurse) \& A (patient) format. We hope that the coding scheme established in this study can aid future research in refining this category with finer granularity, based on specific intervention theories.

\section*{Ethical Considerations}
\label{Ethics}

We used the NLM Scrubber offered by NIH to produce HIPAA-compliant deidentified health information for scientific use, including dates, and places. Two independent annotators evaluated the NLM-Scrubber on the dataset to make sure no events or other people in patients’ posts can allow patients to be traceable. The de-identified version of our data will be shared with researchers upon request who have completed an ethical review from their institution and a data request application form from us.

Since the domain of our dataset is specific, the models trained on our dataset may exhibit subtle biases on out-of-domain data. Further, pre-trained models that we use in our work have been shown to exhibit biases \cite{Li2021OnRA}. We hope future researchers could use these models with caution regarding the biases that these pre-trained models have.

The long-term goal of our work is to \textit{aid} healthcare providers to quickly identify poorly engaged patients to allocate their energy and resources to provide in-time support. Models trained on our data should not be deployed in the real world without human supervision because, despite the potential of transformer models, they cannot be relied on completely in sensitive medical scenarios.

\section*{Acknowledgements}
\label{Ethics}
The randomized clinical trial (WRITE Symptoms intervention) was supported by a grant from the National Institutes of Health, National
Institute of Nursing Research NIH-NINR R01NR010735-NRG Oncology
GOG-259 (H.S.D., PI) as well as NIH-NCI grants to NRG
Oncology (U10CA180822), NRG Operations (U10CA180868), and
UG1CA189867 (NCORP). This study is supported by the Eta Chapter Sigma Theta Tau International Honor Society of Nursing Research Award (Y.W., PI) and the Margaret E. Wilkes Scholarship Fund Award (Y.W., PI). We extend our gratitude to Hannah Bergquist for her invaluable contributions to the annotation of the asynchronous message boards. Furthermore, we are immensely grateful to all the women who participated in this trial, as their involvement has been instrumental to the success of this study.


%



\bibliography{anthology,custom}
\bibliographystyle{acl_natbib}

\appendix


\newpage
\section{Appendix: Mapping of Engagement Categories}
\label{appendixA}
\begin{figure}[!h]
    \centering
    \includegraphics[width=.45\textwidth]{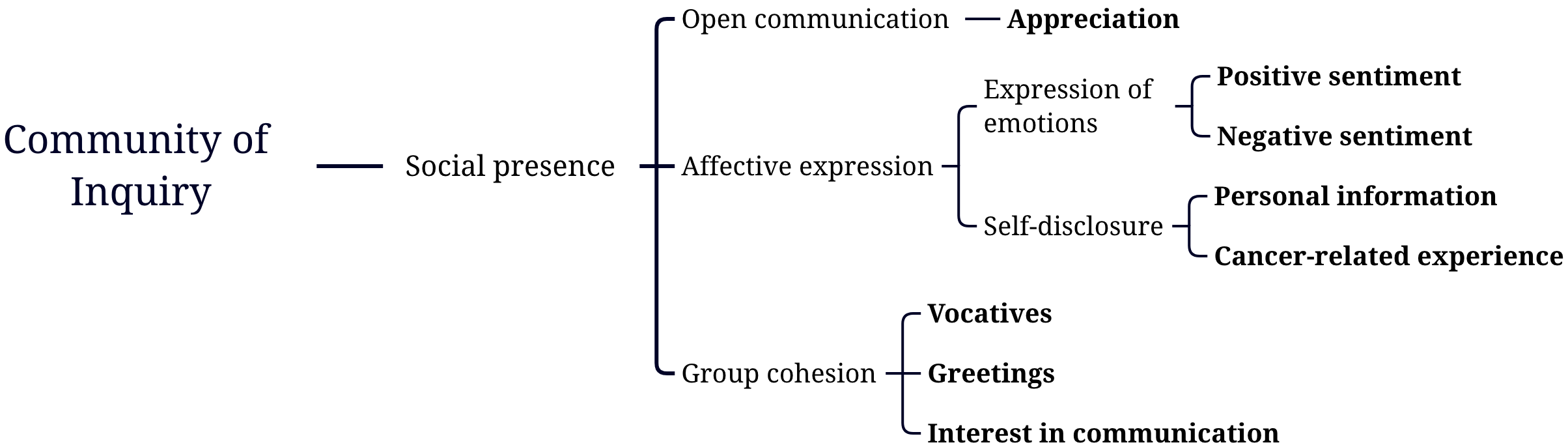}
    \caption{Mapping Community of Inquiry framework \cite{Swan2009ACA} to socio-affective engagement categories.}
    \label{fig:socio-map}
\end{figure}
\begin{figure}[!h]
    \centering
    \includegraphics[width=.45\textwidth]{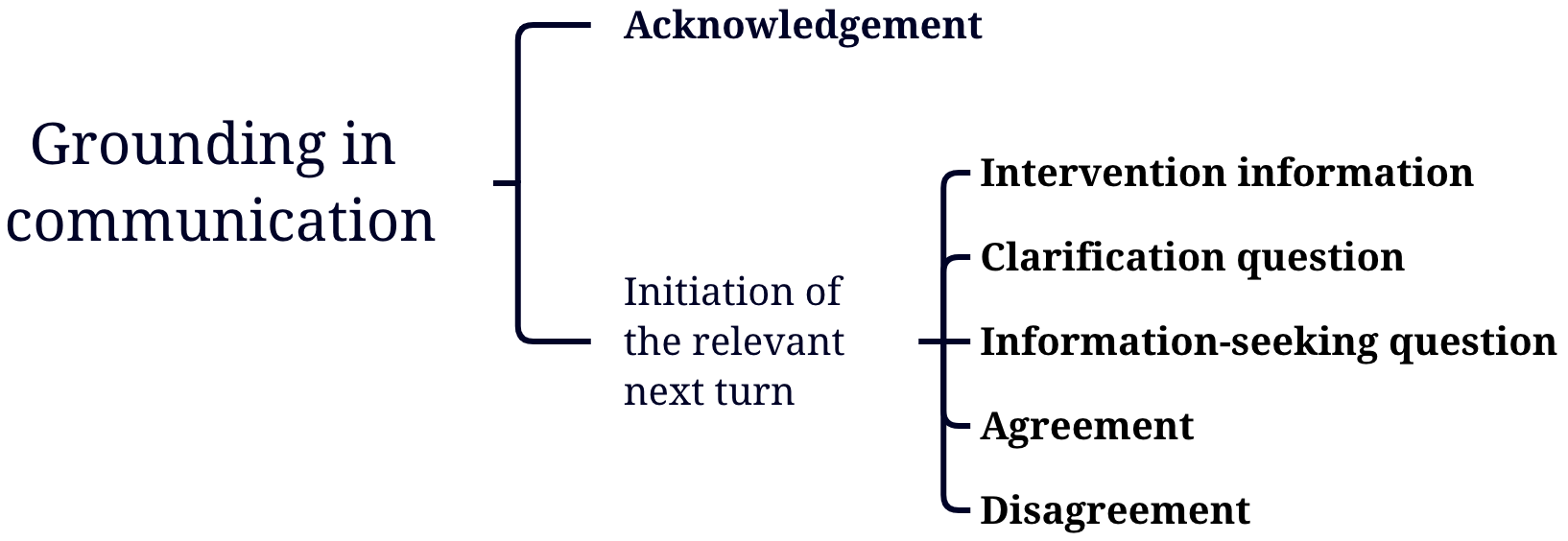}
   \caption{Mapping grounding in communication framework \cite{Clark1991GroundingIC} to cognitive engagement categories.}
    \label{fig:cog-map}
\end{figure}
\section {Appendix: Description of Intervention Process and Coding Scheme}

\label{appendixB}

% ...
\begin{figure*}[!h]
    \centering
    \includegraphics[width=0.9\textwidth]{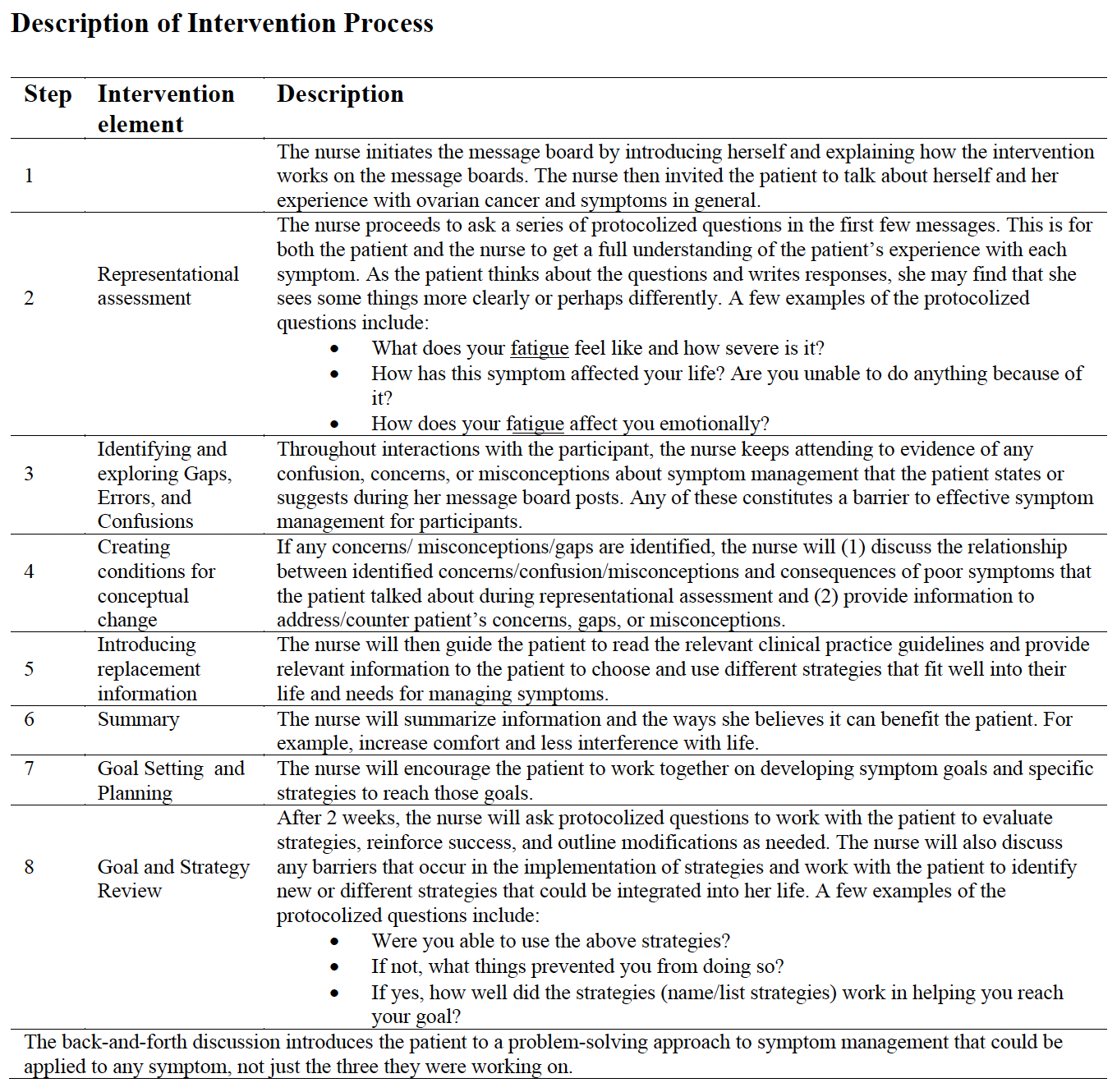}
\end{figure*}


\begin{figure*}[!h]
    \centering
    \includegraphics[width=0.9\textwidth]{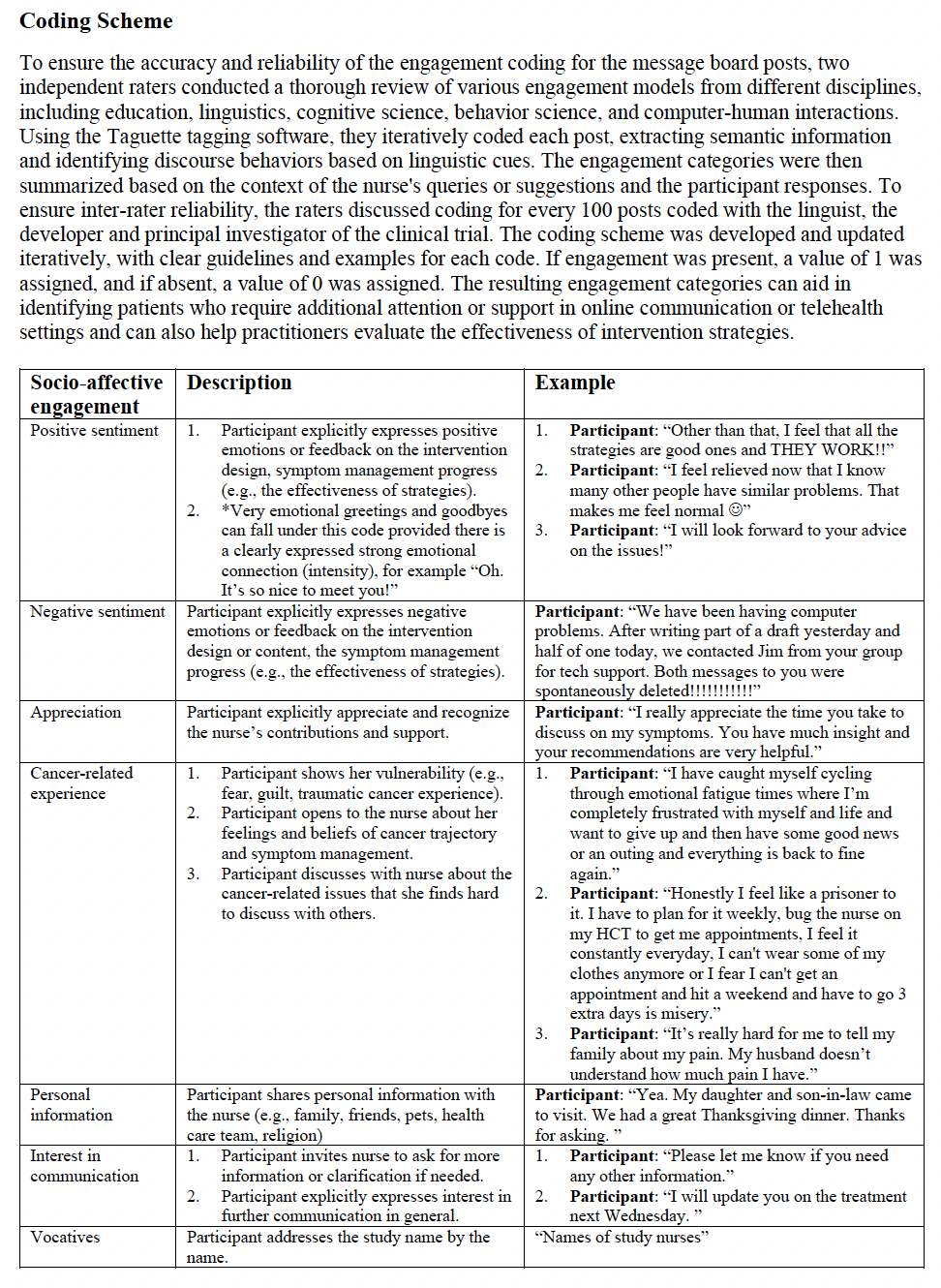}
\end{figure*}

\label{appendixC}
\begin{figure*}[!h]
    \centering
    \includegraphics[width=0.9\textwidth]{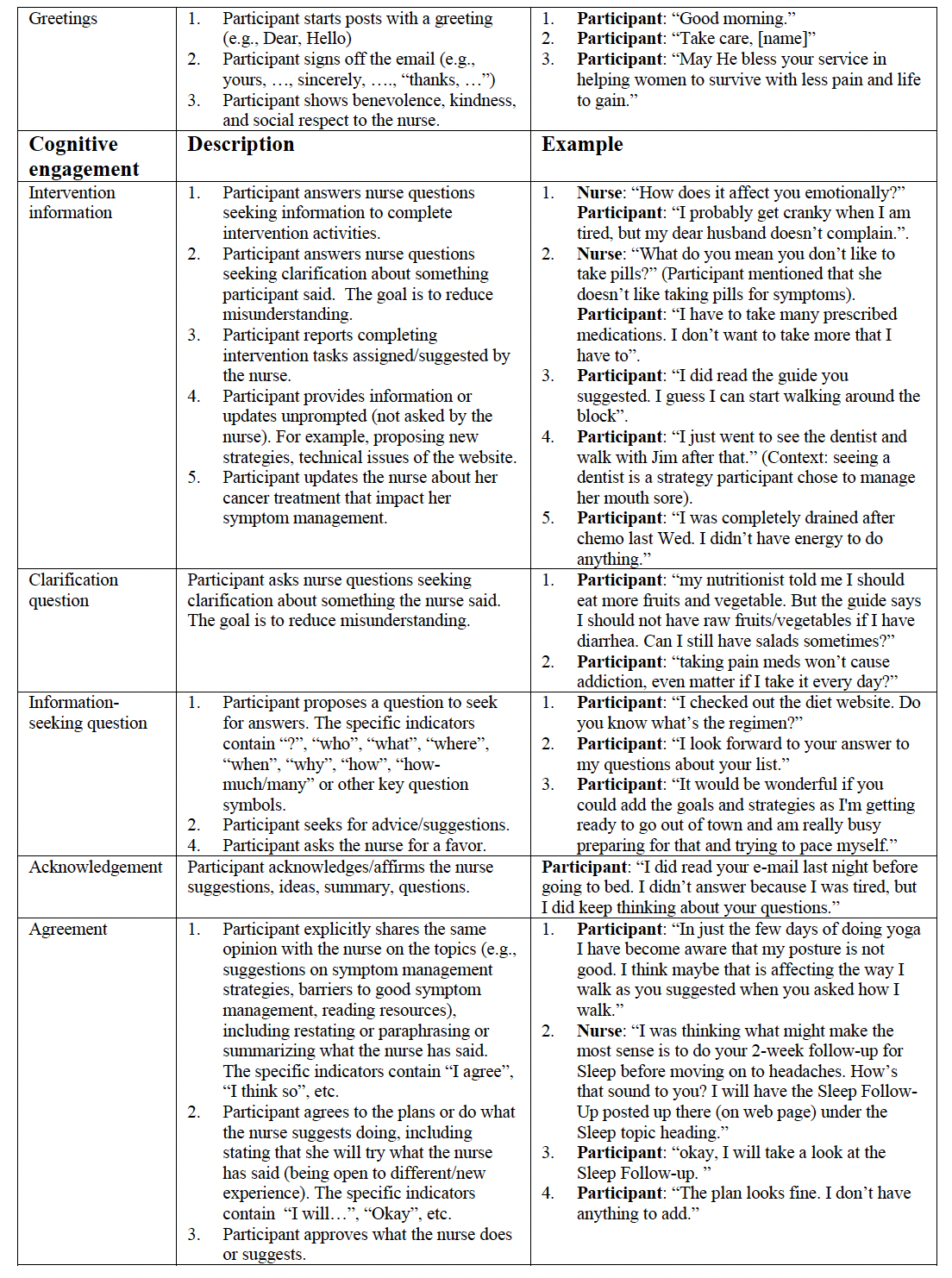}
\end{figure*}

\begin{figure*}[!h]
    \centering
    \includegraphics[width=0.9\textwidth]{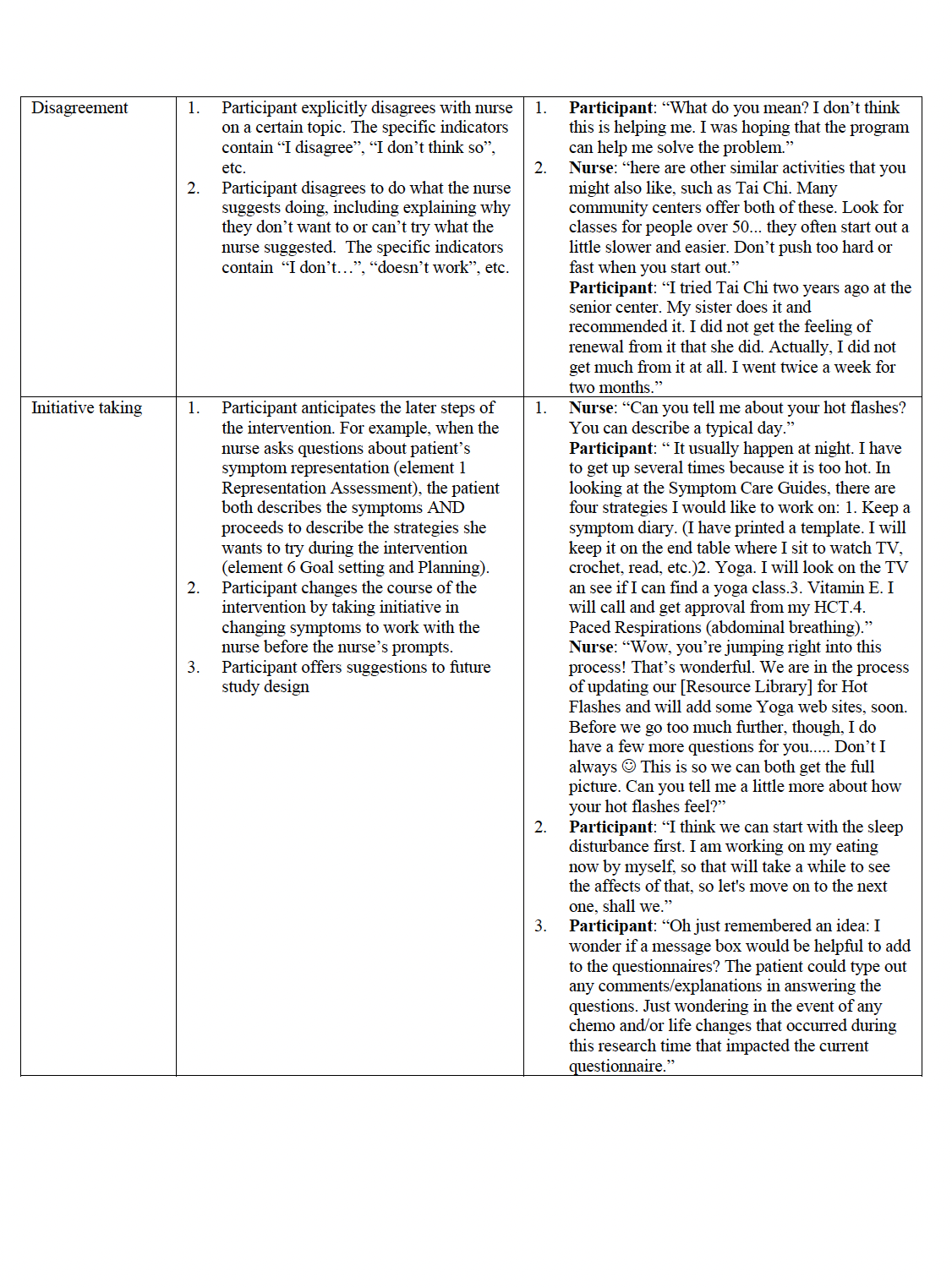}
\end{figure*}




\label{sec:appendix}


\end{document}